\documentclass{article} 
\usepackage{iclr2023_conference,times}

\usepackage{amsmath,amsfonts,bm}

\def\eqref#1{equation~\ref{#1}}

\def\1{\bm{1}}

\DeclareMathAlphabet{\mathsfit}{\encodingdefault}{\sfdefault}{m}{sl}
\SetMathAlphabet{\mathsfit}{bold}{\encodingdefault}{\sfdefault}{bx}{n}

\usepackage{hyperref}
\usepackage{url}

\usepackage{natbib}
\usepackage{graphicx}
\usepackage{amsmath}
\usepackage{amssymb}
\usepackage{mathtools}
\usepackage{amsthm}
\usepackage{booktabs}
\usepackage{algorithm,algorithmic}

\theoremstyle{plain}
\newtheorem{theorem}{Theorem}[section]

\theoremstyle{definition}

\theoremstyle{remark}

\title{LatentAugment: Dynamically Optimized Latent Probabilities of Data Augmentation}

\iclrfinalcopy
\author{Koichi~Kuriyama \\
Division of Natural Resource Economics, Kyoto University, Kyoto, Japan\\
  Kyoto University\\
  Kyoto, Japan 606-8502 \\
  \texttt{kuriyama.koichi.8w@kyoto-u.ac.jp} 
}

\begin{document}

\maketitle

\begin{abstract}
Although data augmentation is a powerful technique for improving the performance of
image classification tasks, it is difficult to
identify the best augmentation policy. The optimal augmentation policy,
which is the latent variable, cannot be directly observed. To address this
problem, this study proposes \textit{LatentAugment}, which estimates the latent probability of optimal
augmentation. The proposed method is appealing in that it can dynamically
optimize the augmentation strategies for each input and model parameter in
learning iterations. Theoretical analysis shows that LatentAugment is a
general model that includes other augmentation methods as special cases, and
it is simple and computationally efficient in comparison with existing
augmentation methods. Experimental results show that the proposed
LatentAugment has higher test accuracy than previous
augmentation methods on the CIFAR-10, CIFAR-100, SVHN, and ImageNet
datasets. 

\end{abstract}

Data augmentation is a widely used technique for generating additional data
to improve the performance of computer vision tasks \citep{shorten2019survey}.
Although data augmentation performs well in
experimental studies, designing data augmentations requires human expertise
with prior knowledge of the dataset, and it is often difficult to transfer
the augmentation strategies across different datasets \citep{krizhevsky2012imagenet}. 
Recent studies on data augmentation consider an
automated design process of searching for augmentation strategies from a
dataset. For example, AutoAugment, proposed by \citet{cubuk2018auto}, uses
reinforcement learning to automatically explore data augmentation policies
using smaller network models and reduced datasets. Although AutoAugment
shows great improvement on image classification tasks of different datasets,
it requires thousands of GPU hours to search for augmentation strategies.
Furthermore, the data augmentation operations optimized for reduced datasets
using smaller network models may not be optimal for full datasets using
larger network models.

To address this problem, this study proposes \textbf{LatentAugment}, which
estimates the latent probability of the optimal augmentation customized to each
input image and network model. There is no doubt that an optimal
augmentation policy exists for each input image using a specific network
model. However, the optimal augmentation policy, which is a latent variable,
cannot be directly observed. Although a latent variable itself cannot be
observed , we can estimate the probability of the latent variable being the
optimal augmentation policy. LatentAugment applies Bayes' rule, to estimate
the conditional probability of the augmentation policy, given the
input data and network parameters.

\begin{figure*}[t]
\begin{center}
\centerline{\includegraphics[width=140mm]{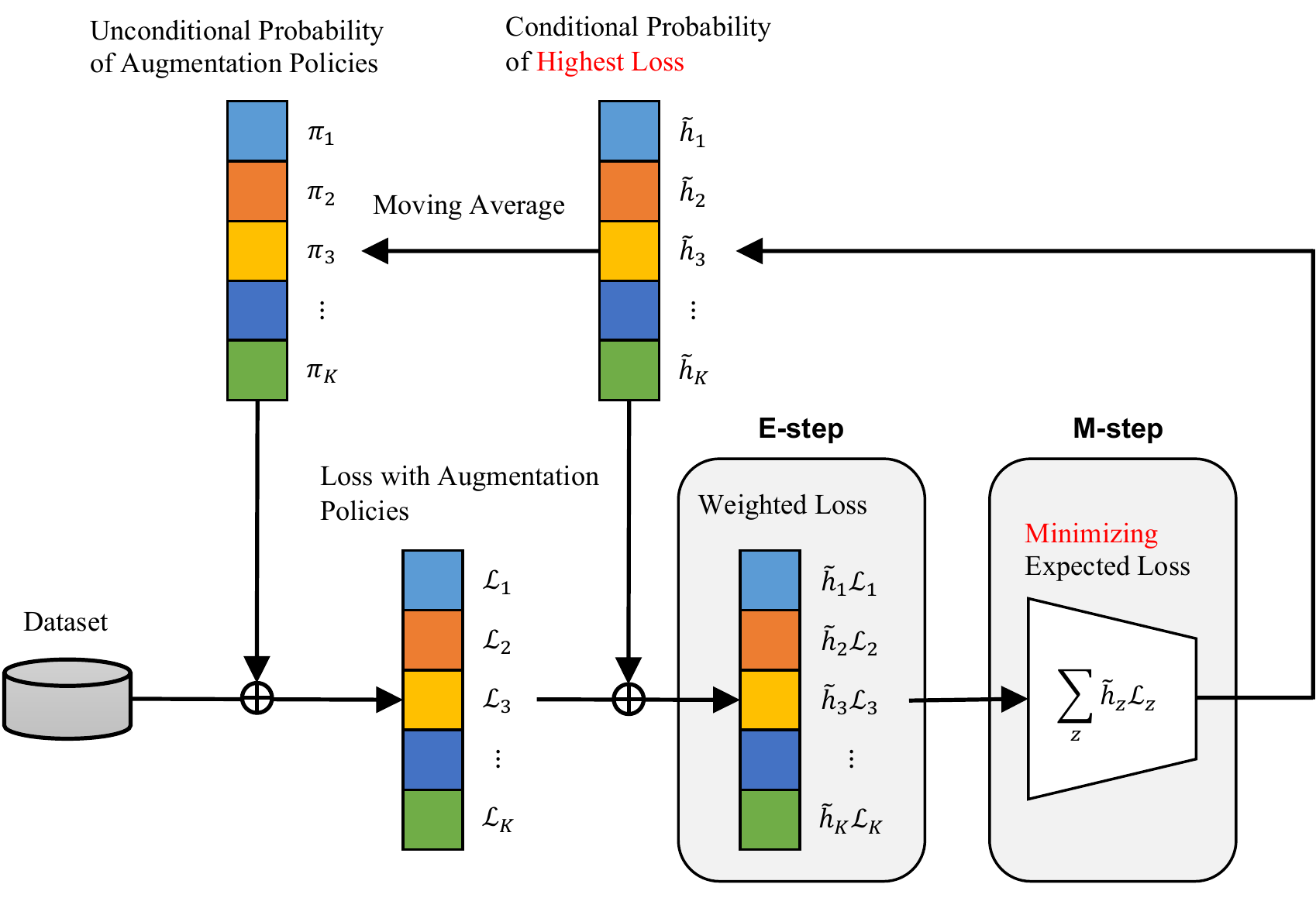}}
\caption{An overview of the proposed LatentAugment. The loss functions with 
augmentation policies are calculated using the input data and the 
unconditional probability of augmentation policies. The model parameters are 
updated by the EM algorithm. In E-step, the expectation of the weighted loss 
function is calculated using the conditional probability of the highest 
loss. In M-step, the expected loss function is minimized using the standard 
stochastic gradient descent. The conditional probabilities of the highest 
loss are calculated using the loss function with the updated parameters and 
input data. The unconditional probabilities of the augmentation policies are 
generated by the moving average of the conditional probability.}
\label{Fig1}
\end{center}
\end{figure*}

Figure \ref{Fig1} shows the concept of the proposed latent augmentation method.
Following the Bayesian data augmentation proposed by \citet{tran2017bayesian},
LatentAugment uses the expectation-maximization (EM) algorithm to update the
model parameters. In the expectation (E)-step, the expectation of the
weighted loss function is calculated using the conditional probability of
the latent augmentation policies. In the maximization (M)-step, the expected
loss function is minimized using the standard stochastic gradient descent.
The conditional probabilities of the highest loss function with the
augmentation policy were calculated using the loss function with
the updated parameters and input data. The unconditional probabilities of
the augmentation policies were generated using the moving average of the
conditional probabilities. Note that the conditional probabilities of the
latent augmentation policies are dynamically optimized for the input and
updated model parameters in the iterations of the EM algorithm.

The contribution of this study can be summarized as follows:

\begin{itemize}
\item It provides a theoretical model for LatentAugment. This study shows that LatentAugment can dynamically optimize the augmentation methods for each input and model parameter in the learning iterations by calculating the conditional probabilities of the latent augmentation policies. Furthermore, it shows that LatentAugment is a general augmentation model that includes other augmentation methods, such as Adversarial AutoAugment \citep{zhang2019adversarial} and uncertainty-based sampling \citep{wu2020generalization}, as special cases.
\item LatentAugment is simple and computationally efficient. It does not require the augmentation policies to be searched before training. Adversarial AutoAugment proposes the application of a generative adversarial network (GAN) \citep{goodfellow2014generative} to solve the maximization of the minimum loss function, which requires an additional training cost for the adversarial network. In contrast, the proposed LatentAugment can solve this problem using the simple stochastic gradient descent algorithm without an adversarial network.
\item Experimental results show that the proposed LatentAugment can improve the test accuracy for the CIFAR-10, CIFAR-100, SVHN, and ImageNet datasets. For example, a test accuracy of 98.72{\%} was achieved with the PyramidNet$+$ShakeDrop \citep{han2017deep, yamada2018shakedrop} on CIFAR-10, which is a significantly better performance compared to previous augmentation methods.
\end{itemize}

\section{Related Works}
\label{sec:related}
Several studies have been conducted on data augmentation methods in the
literature on machine learning. \citet{shorten2019survey} provided a
comprehensive review of image data augmentation. Recent studies have
attempted to automatically identify data augmentation methods. Smart
augmentation \citep{lemley2017smart} merges two or more samples from the same
class to improve the generalization of a target network. AutoAugment (AA)
\citep{cubuk2018auto} applies a recurrent neural network (RNN) as a sample
controller to search for the best data augmentation policy using small proxy
tasks of randomly drawn images from the training dataset. After identifying
the best policy, fixed policies are applied to the training dataset. 
Population-based augmentation (PBA) \citep{ho2019population} generates dynamic
augmentation policy schedules instead of a fixed augmentation policy.
RandAugment (RA) \citep{cubuk2019rand} has a significantly reduced search
space and allows training on the target task without a separate proxy task.
Fast AutoAugment (Fast AA) \citep{lim2019fast} determines the best augmentation
policy using a more efficient search strategy based on density matching.
Faster AA \citep{hataya2019faster} uses a differentiable policy search pipeline
for data augmentation, which is much faster than previous methods.
DADA \citep{li2020differentiable} also  reduces the cost of policy search using
a differentiable optimization problem via Gumbel-Softmax, while DeepAA \citep{zheng2022deepaa}
uses a multi-layer data augmentation pipeline.

Adversarial AutoAugment (AdvAA) \citep{zhang2019adversarial} applies an adversarial
network to generate data augmentation. While the training network minimizes
the loss, the adversarial network maximizes training loss. Uncertainty-Bases
Sampling (UBS) \citep{wu2020generalization} generates data augmentation of the highest
loss without an adversarial network. As shown in the next section, the loss functions of AdvAA and
UBS can be regarded as special cases of LatentAugment proposed in this
study. MetaAugment \citep{zhou2020meta} uses an additional augmentation policy network to 
minimize the weighted losses of augmented training images. DHA \citep{zhou2021dha} 
uses super and child networks to achieve joint optimization of the data augmentation policy, 
hyper-parameter, and architecture. In contrast, the proposed LatentAugment does not 
require any additional network for searching the augmentation policy.

The best augmentation policies are the latent variables that cannot be
observed. The expectation-maximization (EM) algorithm was proposed to
analyze latent variables \citep{dempster1977maximum, mclachlan2007algorithm, ng2012algorithm}.
The EM algorithm estimates parameters using an
iterative process of expectation and minimization of the loss function.
However, when the dataset is large, it might be difficult to calculate the
expectation or minimization of the full dataset. To address the difficulty
of working with large datasets, some approaches have been proposed,
including the generalized EM algorithm \citep{dempster1977maximum}, Monte Carlo
EM algorithm \citep{wei1990monte, tanner1991tools}, stochastic EM algorithm
\citep{nielsen2000stochastic}, and generalized Monte Carlo EM \citep{tran2017bayesian}. For
application to data augmentation, Bayesian data augmentation \citep{tran2017bayesian} 
estimates the parameters using the EM algorithm to generate data
augmentation using the Bayesian approach. Bayesian data augmentation
requires an adversarial network, whereas LatentAugment does not use an
adversarial network. Nevertheless, Bayesian data augmentation is mostly
related to this study in the application of the EM algorithm to data
augmentation.

\section{Method}
Consider a classification task with $C$ categories for the $N$ training data
points $X=\{x_{1},x_{2},\cdots ,x_{N} \}$ and labels $Y=\{y_{1},y_{2},\cdots
,y_{N}\}$. Let $P\left( y\thinspace\vert\thinspace {x,\theta }\right) $ denote
the predicted probability of the output $y$, given the input $x$ and the parameter $\theta$.
Consider that each input is transformed using random data augmentation. Let
$\mathbb{S}=\left\{ 1,\cdots ,S \right\}$ be the set of augmentation policies, and
$z^{\ast }(x,\theta )$ be the optimal augmentation policy for the input given
the parameter. We cannot directly observe the optimal policy; therefore,
$z^{\ast }(x,\theta )$ is the latent variable. Let $\pi_{z}$ be the
unconditional probability that the augmentation policy $z$ is applied to the
input data. The loss function using the augmented data can be written as
$\mathcal{L}\left( \Theta \right)=-\mathbb{E}_{\left( x,y \right)\sim \left( X,Y
\right)}\mathrm{log}\left( \sum\limits_{z\in \mathbb{S}} {\pi_{z}P\left(
y\thinspace\vert\thinspace {o_{z}\left( x \right),\theta }\right) } \right)$,
where $o_{z}(x)$ denotes the augmented data using the augmentation policy
$z$, $\Theta = \{\theta, \pi\}$ and $\pi = \{\pi_1,\cdots,\pi_{S}\}$.

\subsection{Generalized EM algorithm}

The loss function with the latent variable can be minimized using the
expectation--maximization (EM) algorithm. The EM algorithm is an iterative
procedure used to compute the maximum likelihood estimate in the presence of
latent variables \citep{ng2012algorithm}. In the E-step, the expected loss function calculated. In
the M-step, the parameter is updated by minimizing the expected loss
function. Let $\Theta^{(t)} = \{\theta^{(t)}, \pi^{(t)}\}$ be the parameter and the unconditional probability at iteration $t$ and
$h_{z}^{(t)}(x,y,\Theta^{\left( t \right)},\mathbb{S})$ be the conditional probability
of the augmentation probability of policy $z$ for the individual data point
$x$ given the label $y$ at iteration $t$. Applying Bayes' rule, we can
calculate the conditional probability \citep{mclachlan2007algorithm}:
\[
h_{z}^{(t)}(x,y,\Theta^{\left( t \right)},\mathbb{S})=\frac{\pi_{z}^{(t)}P\left(
y\thinspace\vert\thinspace {o_{z}\left( x \right),\theta^{(t)}}\right)
}{\sum\limits_{z\in \mathbb{S}} {\pi_{z}^{(t)}P\left( y\thinspace\vert\thinspace
{o_{z}\left( x \right),\theta^{(t)}}\right) } }.
\]
Using $h_{z}^{(t)}$, as shown in \citet{ng2012algorithm}, the expected loss function $\mathcal{E}\left( \Theta \vert \Theta
^{\left( t \right)} \right)$ can be written as:
\begin{equation}
\label{eq:expected_loss}
\mathcal{E}\left( \Theta \vert \Theta^{\left( t \right)} \right)=-\mathbb{E}_{\left( x,y
\right)\sim \left( X,Y \right)}
\left[ \sum\limits_{z\in \mathbb{S}} {h_{z}^{(t)}\log
\left( \pi_{z} \right)}\right.
\left.+\sum\limits_{z\in \mathbb{S}} {h_{z}^{(t)}\log \left(
P\left( y\thinspace\vert\thinspace {o_{z}\left( x \right),\theta }\right)
\right)} \right].
\end{equation}

In the M-step, the parameter $\theta $ and the unconditional probability
$\pi_{z}$ were estimated by minimizing the expected loss function given the
conditional probability $h_{z}^{(t)}$. If solving the
minimization problem of $\mathcal{E}\left( \Theta \vert \Theta^{\left( t \right)}
\right)$ proves difficult, the generalized EM algorithm proposed by \citet{dempster1977maximum} can be used to estimate $\Theta^{(t+1)}$, where $\mathcal{E}\left(
\Theta^{(t+1)}\vert \Theta^{\left( t \right)} \right)<\mathcal{E}\left( \Theta \vert
\Theta^{\left( t \right)} \right)$. 

Calculation of $\mathcal{E}\left( \Theta \vert \Theta^{(t)} \right)$ requires the
expectation of possible augmentation policies. When the number of
augmentation policies $\mathbb{S}$ is large, the computational burden of $\mathcal{E}\left(
\Theta \vert \Theta^{(t)} \right)$ cannot be neglected. Alternatively, the
subset $\mathbb{K}$ which is randomly drawn from the full set $\mathbb{S}$, of the
augmentation policies, can be used. Then, the conditional probability
$h_{z}^{(t)}$ can be written as
\begin{equation}
\label{eq:hz_k}
h_{z}^{\left( t \right)}\left( x,y,\Theta^{\left( t \right)},\mathbb{K}
\right)=\frac{\pi_{z}^{(t)}P\left( y\thinspace\vert\thinspace {o_{z}\left( x
\right),\theta^{\left( t \right)}}\right) }{\sum\limits_{z\in \mathbb{K}} {\pi
_{z}^{(t)}P\left( y\thinspace\vert\thinspace {o_{z}\left( x \right),\theta
^{\left( t \right)}}\right) } }.
\end{equation}

As shown in the Appendix (\ref{sec:A1}), if the subset $\mathbb{K}$ is generated using simple random
draws from the full set $\mathbb{S}$, the expected loss function using the subset is
equal to that obtained using the full set.

\subsection{Latent Augmentation Policy}

The generalized EM algorithm can estimate the parameter by minimizing the
expected loss function using latent variables. However, this may cause an
overfitting problem. Following AdvAA, an augmentation policy is applied to
maximize the loss function using a harder augmentation policy. Let $\mathcal{L}_{z}^{(t)}=
-\log \left( \pi_{z}^{(t)}\cdot P\left( y\thinspace\vert\thinspace {o_{z}\left( x
\right),\theta^{(t)}}\right) \right)$ be the the contribution to the loss function for input $\left( x,y \right)$,
using the augmentation policy $z$ at iteration $t$. Consider the conditional probability of the latent 
augmentation policy $\Tilde{h}^{(t)}_z$ such that the augmentation policy $z$ has the highest loss  
in the set of $\mathbb{K}$, using the softmin function:

\begin{equation}
\Tilde{h}^{(t)}_z  = \Pr[\mathcal{L}^{(t)}_z\ge \mathcal{L}^{(t)}_k, \forall k\in \mathbb{K}] 
= \Pr[h^{(t)}_z\le h^{(t)}_k, \forall k\in \mathbb{K}]=\frac{\exp(-h^{(t)}_z /\sigma)}{\sum_{k\in \mathbb{K}} \exp(-h^{(t)}_k /\sigma)},
\label{eq:h_tilde}
\end{equation}

where $\sigma $ is the inverse scale parameter. Note that, from the definition of $\mathcal{L}_z$, the probability of minimum $h_z$ is equal to the one of maximum $\mathcal{L}_z $. Thus, the softmin function is related to the goal of the LatentAugment, maximization of minimum loss. The proposed LatentAugment can be implemented by the EM algorithm with $\Tilde{h}^{(t)}_z$. In the E-step, the LatentAugment 
calculates the expected loss function weighted by the probability of the minimum conditional 
probability $\Tilde{h}_{z}^{\left( t \right)}$, instead of $h_{z}^{\left( t \right)}$:
\begin{equation}
\label{eq:E_step_tilde}
\Tilde{\mathcal{E}}\left( \Theta \vert \Theta^{(t)} \right)=-\mathbb{E}_{\left( x,y \right)\sim
\left( X,Y \right)}\mathbb{E}_{\mathbb{K} \sim \mathbb{S}}\left[ \sum\limits_{z\in \mathbb{K}}
{\Tilde{h}_{z}^{\left( t \right)}\log \left( \pi_{z} \right)}\right.
\left.+\sum\limits_{z\in \mathbb{K}} {\Tilde{h}_{z}^{\left( t \right)}\log \left( P\left(
y\thinspace\vert\thinspace {o_{z}\left( x \right),\theta }\right) \right)}
\right].
\end{equation}

In the M-step, the parameter is updated by minimizing the expected loss
function with fixed $\Tilde{h}_z^{(t)}$:
\begin{equation}
\label{eq:M_step_tilde}
\theta^{(t+1)}=\theta^{(t)}-\eta \mathrm{\nabla }_{\theta }\Tilde{\mathcal{E}}\left(
\Theta \vert \Theta^{(t)} \right),
\pi_{z}^{(t+1)}=\mathrm{Moving\, Average\, of}\, \frac{\mathbb{E}_{\left( x,y
\right)\sim \left( X_{B},Y_{B} \right)}\Tilde{h}_{z}^{\left( t
\right)}}{\mathbb{E}_{\left( x,y \right)\sim \left( X_{B},Y_{B}
\right)}\sum\limits_{z\in \mathbb{K}} \Tilde{h}_z^{(t)} },
\end{equation}
where $\left( X_{B},Y_{B} \right)$ is the mini-batch of the input data. This
process is iterated until convergence is achieved. The estimation procedure
with LatentAugment is summarized as Algorithm \ref{alg:LA_alg}.

\begin{algorithm}[tb]
   \caption{LatentAugment}
   \label{alg:LA_alg}
\begin{algorithmic}
   \STATE {\bfseries Input:} $\left( X,Y \right)$: dataset
   \STATE {\bfseries Require:} $B$: the number of mini-batch, $\mathbb{S}$: the full set of 
augmentation policies, $S$: the size of augmentation policies, and $\sigma $: the inverse scale.
   \STATE {\bfseries Initialize:}  $\pi_{z}=1/S$, for $z=\{1,\mathellipsis ,S\}$. Initialize the network parameter 
$\theta^{(0)}$.
   \FOR{$t=1,\mathellipsis ,B$}
   \STATE Randomly draw the subset $\mathbb{K}$ from $\mathbb{S}$.
   \STATE Calculate $\tilde{h}_{z}^{(t)}$ using equation (\ref{eq:h_tilde}).
   \STATE E-step: Calculate $\Tilde{\mathcal{E}}$ using equation (\ref{eq:E_step_tilde}) 
   \STATE M-step: Update the parameter $\theta^{(t)}$ and $\pi^{(t)}$ using equation (\ref{eq:M_step_tilde}) 
  
   \ENDFOR
   \STATE \textbf{Return:} $\theta^{(B)}$ and $\pi^{(B)}$
\end{algorithmic}
\end{algorithm}

\subsection{Advantages of the LatentAugment}

The proposed LatentAugment has following advantages over the existing
augmentation methods:

\begin{enumerate}
\item \textbf{The weighted augmentation policies are optimized for the individual input.}
Most recent studies, such as AA, use randomly drawn policies; however, they
do not apply policies that are appropriate for each input data. On the other
hand, LatentAugment utilizes randomly drawn policies customized for each
input by calculating conditional probabilities for the given input.

\item \textbf{It provides a closed-form solution for the probability of optimal augment polices.}
LatentAugment can estimate the unconditional probability ($\pi_z$) of optimal augment policies using
a closed-form solution (\ref{eq:M_step_tilde}) of the loss minimization. 
Thus, LatentAugment does not involve the additional cost of searching for these policies.

\begin{table}[b]
\centering
\caption{Comparing the training cost and test accuracy between proposed LatentAugment (LA) with 
RandAugment (RA \citep{cubuk2019rand}), Adversarial AutoAugment (AdvAA \citep{zhang2019adversarial}), and Uncertainty-Based 
Sampling (UBS \citep{wu2020generalization}) using the Wide-ResNet 28-10 model on CIFAR-10. Training cost of required GPU hours is reported relative to RA.}
\label{tab1}
\small
\begin{tabular}{ccccccc}
\toprule
& 
RA& 
AdvAA& 
UBS& 
LA ($K=2)$&
LA ($K=4)$&
LA ($K=6)$ \\
\midrule
Training cost& 
1& 
8& 
1.5& 
1.9&
3.3&
4.7 \\
Test accuracy&
95.8\%&
98.1\%&
97.9\%&
98.0\%&
98.2\%&
98.3\% \\
\bottomrule
\end{tabular}
\vskip -0.1in
\end{table}

\item \textbf{It is simple and computationally efficient.}
AdvAA proposed the use of GAN to solve for the maximization of the minimum
loss function, which requires additional training costs for the adversarial
network. In contrast, the proposed LatentAugment can solve the max-min
problem using the conditional probability ($\Tilde{h}_{z}^{\left( t
\right)})$ of the highest loss without an adversarial network. LatentAugment can be solved using
a simple stochastic gradient descent algorithm. Table \ref{tab1} compares the training cost of required GPU hours
between proposed LatentAugment and other methods.

\item \textbf{It is a general model which includes other augmentation methods.}
The proposed LatentAugment is a general augmentation method that includes
other methods such as UBS and AdvAA. Following theorem shows that the UBS with a single data point
and AdvAA could be considered to be special cases of LatentAugment:

\vskip.5\baselineskip
\end{enumerate}

\begin{theorem}
\label{thm:theorem1}
(Special Case of LatentAugment). Assume that the unconditional probabilities for all augmentation policies are the same ($\pi_z=1/S, \forall z$). If the inverse scale parameter $\sigma \to 0$, the gradient of expected loss function of the LatentAugment can be equal to the one of UBS. If $\sigma \to \infty$, the gradient of expected loss function of the adversarial network with LatentAugment is equal to the one of AdvAA.

\end{theorem}

The proof can be found in Appendix \ref{sec:A2}. Note that the the gradient of expected loss function of LatentAugment with $\sigma \to \infty$ can be equivalent to the one of AdvAA. However, it means that LatentAugment can not maximize minimum loss without an additional network, thus the advantages in efficiency of LatentAugment will be lost.

\section{Experiments}
\label{sec:experiments}

\subsection{Experiment Setting}
This section describes the experiments investigating the performance of the
proposed LatentAugment using the CIFAR-10 and CIFAR-100 \citep{krizhevsky2009learning}, 
SVHN \citep{netzer2011reading}, and ImageNet \citep{russakovsky2015imagenet} datasets. In these experiments, the network models are Resnet-50 \citep{he2016deep}, Wide-ResNet 40-2 and Wide-ResNet 28-10 \citep{Zagoruyko2016}, Shake-Shake 26 2\texttimes 32d, 26 2\texttimes 96d,
and 26 2\texttimes 112d \citep{gastaldi2017shake}, and PyramidNet with ShakeDrop
with a depth of 272 and an alpha of 200 \citep{han2017deep, yamada2018shakedrop}.
Table \ref{tab_a1} in the Appendix \ref{sec:A4} provides the hyperparameters. All the
hyperparameters of the network models are the same as those used in AA
\citep{cubuk2018auto}, Fast AA \citep{lim2019fast}, and PBA \citep{ho2019population}.
A cosine learning decay with one annealing cycle was applied to all models.

As proposed by \citet{wu2020generalization}, we use 16 transformations: AutoContrast,
Brightness, Color, Contrast, Cutout, Equalize, Invert, Mixup, 
Posterize, Rotate, Sharpness, ShearX, ShearY, Solarize, TranslateX,
and TranslateY.  \citet{shorten2019survey} described these
transformations. As in AA, augmentation policies are generated by the
combination of the two transformations. The size of the policy set was
$S=16\times 16=256$. Following AA, we set the magnitude of all the transform operations 
in a moderate range. All values of the magnitude range are same as AA.

The unconditional probability ($\pi_{z})$ was initialized as $1
\mathord{\left/ {\vphantom {1 S}} \right. \kern-\nulldelimiterspace} S$. The
range of the magnitude of each transformation was discretized into 10, which
were randomly drawn from the uniform distribution. The unconditional
probability ($\pi_{z})$ was calculated using the moving average. The length
of the moving average was fixed at 10 iterations in this experiment. It was
difficult to estimate the expected loss function using the full set $\mathbb{S}$ of
augmentation policies because of the computational burden of the large size
$S=256$. Alternatively, subset $\mathbb{K}$ could be used which is randomly drawn
from the full set $\mathbb{S}$. In this experiment, the subset size of the
augmentation policies was set to six ($K=6)$. The inverse scale parameter
$\sigma $ was set to one. The effects of the unconditional probability,
subset size, and inverse scale are discussed in later sections of the paper.

\begin{table*}[t]
\centering
\caption{Test accuracy ({\%}) on CIFAR-10, 
CIFAR-100, SVHN, and ImageNet. All experiments in this study replicate the 
results of Baseline and AutoAugment (AA) \citep{cubuk2018auto}, 
Adversarial AutoAugment (AdvAA) \citep{zhang2019adversarial}, 
 Uncertainty-Based Sampling (UBS) \citep{wu2020generalization}, and MetaAugment (MA) \citep{zhou2020meta}. 
On the proposed LatentAugment (LA), averages of five runs are reported.  
Network models are Wide-ResNet 40-2 and Wide-ResNet 28-10 \citep{Zagoruyko2016}, 
Shake-Shake 26 2\texttimes 32d, 26 2\texttimes 96d, and 26 2\texttimes 112d \citep{gastaldi2017shake}, 
PyramidNet with ShakeDrop \citep{han2017deep, yamada2018shakedrop} and Resnet-50 \citep{he2016deep}.
See text for more details.
}
\vskip 0.15in
\begin{center}
\small
\begin{tabular}{cccccccc}
\toprule
Dataset& 
Model& 
Baseline& 
AA& 
AdvAA& 
UBS&
MA& 
LA \\
\midrule
CIFAR-10& 
WRN40-2& 
94.70& 
96.30& 
-& 
-& 
96.79 &
\textbf{97.27\textpm 0.09} \\
& 
WRN28-10& 
96.13& 
97.32& 
98.10& 
97.89& 
97.76&
\textbf{98.25\textpm 0.08} \\
& 
S-S (26 2x32d)& 
96.45& 
97.53& 
97.64& 
-& 
-&
\textbf{97.68\textpm 0.03} \\
& 
S-S (26 2x96d)& 
97.14& 
98.01& 
98.15& 
98.27& 
98.29&
\textbf{98.42\textpm 0.02} \\
& 
S-S (26 2x112d)& 
97.18& 
98.11& 
98.22& 
-& 
98.28&
\textbf{98.44\textpm 0.02} \\
& 
PyramidNet& 
97.33& 
98.52& 
98.64& 
98.66& 
98.57&
\textbf{98.72\textpm 0.02} \\
\midrule
CIFAR-100& 
WRN40-2& 
74.00& 
79.30& 
-& 
-& 
80.60&
\textbf{80.90\textpm 0.15} \\
& 
WRN28-10& 
81.20& 
82.91& 
84.51& 
84.54& 
83.79&
\textbf{84.98\textpm 0.12} \\
& 
S-S (26 2x96d)& 
82.95& 
85.72& 
85.90 & 
-& 
\textbf{85.97}&
85.88\textpm 0.10 \\
\midrule
SVHN& 
WRN28-10& 
98.50& 
98.93& 
-& 
-& 
-&
\textbf{98.96\textpm 0.01} \\
\midrule
ImageNet& 
ResNet-50 (Top 1)& 
75.30& 
77.63& 
79.40& 
-& 
79.74&
\textbf{80.02\textpm 0.10} \\
&
ResNet-50 (Top 5)& 
92.20& 
93.82& 
94.47& 
-& 
94.64&
\textbf{94.88\textpm 0.05} \\
\bottomrule
\end{tabular}
\label{tab2}
\end{center}
\vskip -0.1in
\end{table*}

\subsection{CIFAR-10 Results}
The CIFAR-10 dataset has a total of 60,000 images, including 50,000 for the
training set and 10,000 for the test set. Each image with a size of 32
\texttimes 32 belongs to one of the 10 classes. The baseline is trained with
standard data augmentation using horizontal flips with 50{\%} probability,
zero-padding, and random crops. The proposed LatentAugment first applies the
baseline preprocessing, then applies LatentAugment using six policies
randomly drawn from 256 policies, and finally applies the Cutout \citep{devries2017improved} or the Cutmix \citep{yun2019cutmix}.

Table \ref{tab2} shows the results of the test accuracy for different network models
using the CIFAR-10 dataset. For all models, the proposed LatentAugment
method achieved a better performance compared to existing augmentation
methods. For example, LatentAugment achieved an improvement of 0.15{\%} and
0.36{\%} compared to AdvAA and UBS on Wide-ResNet 28-10 model, respectively.
The test accuracy of the proposed LatentAugment using PyramidNet$+$ShakeDrop
was 98.72 {\%}, which was 0.08{\%} and 0.06{\%} better than that of AdvAA
and UBS, respectively. To compare with the AA, we tested the proposed model
using the same transformations as AA, which uses the policy set with
SamplePairing \citep{inoue2018data}, instead of Mixup \citep{zhang2017mixup}, and
finally applied the Cutout, instead of Cutmix.
The test accuracies for the Wide-ResNet 40-2 model and Wide-ResNet 28-10 of
the LatentAugment using the same transformation as AA were 96.91\textpm
0.05{\%} and 98.01\textpm 0.05{\%}, respectively (Table \ref{tab_acc_same_aa}). Thus, the proposed method
outperforms AA, even when neither Mixup nor Cutmix were used. Adversarial 
AutoAugment (AdvAA) also applies the same transformations as AA, although 
the subset size of AdvAA is 8. To compare with AdvAA, we tested the model using 
the same subset size. The test accuracy of Wide-ResNet 28-10 of the LatentAugment 
using the same transformations as AA with subset size $K=8$ is 98.16±0.07\%. Thus, 
the proposed method is marginally better than AdvAA even when the same transformations 
and subset sizes are used.  See table \ref{tab_acc_same_aa} and \ref{tab_acc_same_advaa} in Appendix.

\subsection{CIFAR-100 Results}
The CIFAR-100 dataset also has a total of 60,000 images, including 50,000
for the training set and 10,000 for the test set. The number of categories
is 100. The procedure of the baseline and LatentAugment is the same
as that of CIFAR-10.

As for CIFAR-10, the proposed LatentAugment indicated better accuracy than
existing augmentation methods except Shake-Shake (26 2\texttimes 96d), in
which the test accuracy of LatentAugment was slightly lower than that of
AdvAA and MetaAugment (MA).

\subsection{SVHN Results}
The SVHN dataset has 73,257 digit images for the core training set, 531,131
for the additional training set, and 26,032 for the test set. In this
experiment, both core and additional training sets were used. The number of
categories is 10. The baseline was trained using the normalizing
data. The proposed method first applies LatentAugment using six policies,
randomly drawn from 256 policies, then normalizes the data, and finally
applies the cutout with a region size of 20 \texttimes 20 pixels, following
the method proposed by  \citet{devries2017improved}. LatentAugment using
Wide-ResNet 28-10 achieves 0.03{\%} improvement compared to AA.

\subsection{ImageNet Results}
The ImageNet dataset has more than 1.2 million training images, 50,000
validation images, and 100,000 test images. The number of categories is
1,000. Following the AA, baseline augmentation uses the standard
Inception-style pre-processing, including horizontal flips with 50{\%}
probability and random distortions of colors. The proposed LatentAugment
first applies the baseline preprocessing, then applies LatentAugment using
six policies randomly drawn from 256 policies, and finally applies the
Cutmix. The proposed method outperformed previous augmentation studies.

\subsection{Choice of the Subset Size}
\begin{figure}[tb]
\centerline{\includegraphics[width=2.8in]{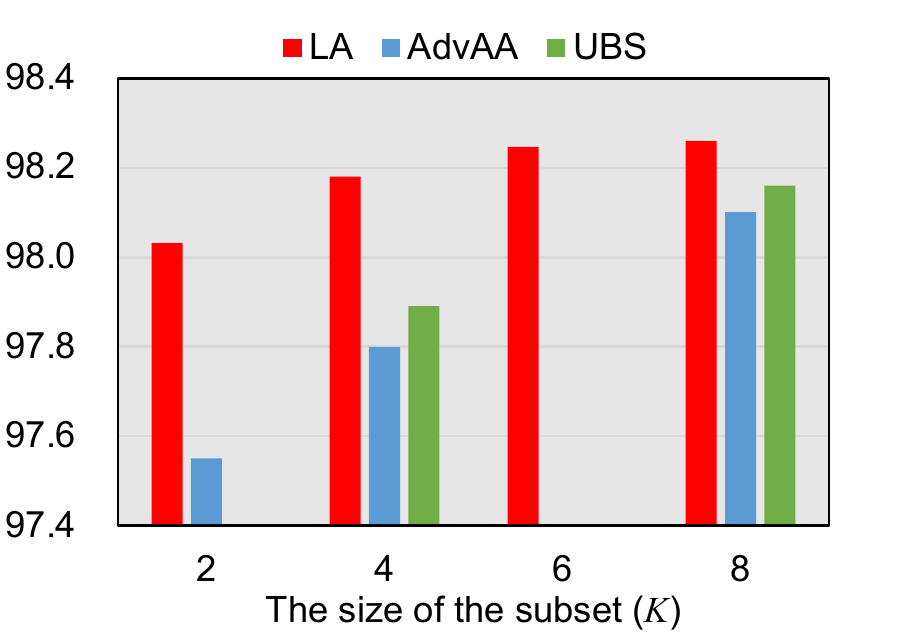}}
\caption{The test accuracies with the 
different size of the subset ($K)$. It shows the test accuracies of 
LatentAugment (LA), Uncertainty-Based Sampling (UBS), Adversarial 
AutoAugment (AdvAA) with the different size of the subset ($K)$ using the 
Wide-ResNet 28-10 model on CIFAR-10. This figure replicates the results of 
UBS from \citet{wu2020generalization} and AdvAA from \citet{zhang2019adversarial}.}
\label{Fig3}
\end{figure}

This experiment used a subset size of $K=6$. To determine the optimal size
of the subset, this study used the Wide-ResNet 28-10 to evaluate the
performance of the proposed LatentAugment with different $K$, where $K\in
\{2,4,6,8\}$. Figure \ref{Fig3} suggests that the test accuracy of the model rapidly
increases up to $K=6$. However, no significant improvement was observed when
$K$ was 8. In contrast, the computational cost increases with $K$.
Therefore, after comparing the computational cost and performance, all the
experiments in this study used $K=6$ for LatentAugment.

Figure  \ref{Fig3} also shows the results of AdvAA and UBS. AdvAA uses instances of 
$K \in \{2,4,8,16,32\}$  for each input example, augmented by adversarial policies. The study
of UBS reports the experimental results using $K=4$ with a single data point
and $K=8$ with four data points for training. This figure suggests that the
proposed LatentAugment is more efficient than AdvAA and UBS, because
LatentAugment with $K=4$ outperforms AdvAA and UBS with $K=8$.

\subsection{The Effects of the Inverse Scale}
\begin{figure}[htbp]
\centerline{\includegraphics[width=2.8in]{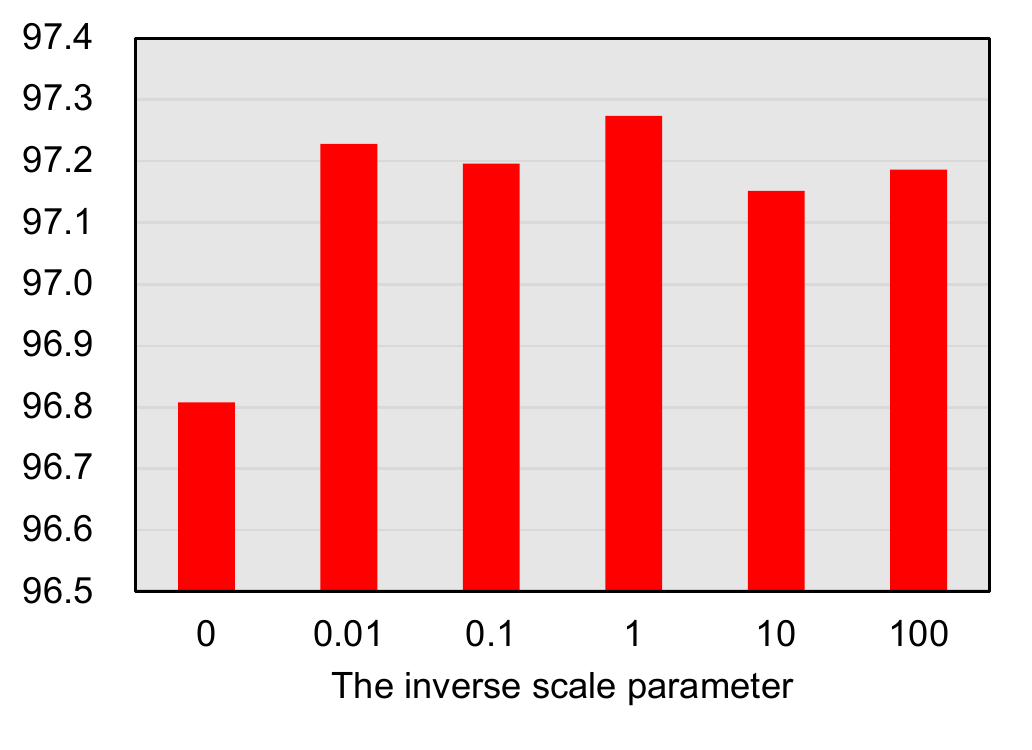}}
\caption{The test accuracies with the different inverse scale 
parameter ($\sigma )$. It shows the test accuracies of LatentAugment with 
the different inverse scale values using the Wide-ResNet 40-2 model on 
CIFAR-10.}
\label{Fig4}
\end{figure}

LatentAugment requires determining the inverse scale parameter $\sigma $
which is assumed to have a value of 1 in the previous section. This section
considers the effects of the inverse scale using different values. 
Figure \ref{Fig4} shows the test accuracy of LatentAugment with
different inverse scale values using the Wide-ResNet 40-2 model on CIFAR-10.
This suggests that the test accuracy is maximum at $\sigma =1$, although the
effect of the inverse scale is weak except for $\sigma \to 0$.

\subsection{The Effects of the Unconditional Probability}
LatentAugment estimates the unconditional probabilities ($\pi_{s})$ as well
as the network parameters ($\theta )$. As shown in Theorem \ref{thm:theorem1}, if the
unconditional probabilities are fixed at the same values, the derivative of
LatentAugment can be reduced to that of adversarial AA or UBS. Table \ref{tab3} shows
the effects of the unconditional probability in LatentAugment using the
Wide-ResNet 40-2 model on CIFAR-10. The cell of (a), where the unconditional
probabilities ($\pi_{z})$ are fixed and the inverse scale parameter
($\sigma )$ is set to 0, is equivalent to UBS with a single data point of
the highest loss. In contrast, the cell of (d), where $\pi_{z}$ can be
estimated and $\sigma =1$, is the test accuracy of the proposed
LatentAugment, which allows variable $\pi_{z}$ and multiple data points for
the expectation of the loss function. This table suggests that an unfixed
$\pi_{z}$ can slightly improve the test accuracy over a fixed $\pi_{z}$.
However, the effect on test accuracy with an fixed $\pi_{z}$ is weaker
than the effect of $\sigma $ set to zero. Thus, for the better performance
of the proposed LatentAugment, using multiple data points for the expected
value of the loss function weighted by the conditional probability of the
highest loss, has a more significant effect on the performance than using the
unfixed unconditional probability.

\begin{table}[tbp]
\begin{center}
\caption{The test accuracies using fixed or unfixed 
unconditional probability with the different inverse scale parameters. 
Averages of five runs are reported.}
\vskip 0.15in
\small
\begin{tabular}{cccc}
\toprule
\raisebox{-1.50ex}[0cm][0cm]{}& 
\multicolumn{2}{c}{
\begin{tabular}{c}
Inverse scale parameter ($\sigma )$
\end{tabular}
} & 
\raisebox{-1.50ex}[0cm][0cm]{Diff.} \\
 & 
$\sigma =0$& 
$\sigma =1$& 
 \\
\midrule
Fixed $\pi_{z}$& 
(a) 96.71\textpm 0.10& 
(b) 97.19\textpm 0.08& 
0.48 \\
Unfixed $\pi_{z}$& 
(c) 96.81\textpm 0.10& 
(d) 97.27\textpm 0.09& 
0.47 \\
\midrule
Diff.& 
0.10& 
0.08& 
 \\
\bottomrule
\end{tabular}
\label{tab3}
\end{center}
\end{table}

\section{Conclusions}
\label{Conclusions}
This study introduces LatentAugment, which estimates the probability of the latent augmentation customized to each input image and network model. The proposed method is appealing in that it can dynamically optimize the augmentation methods for each input and model parameter in learning iterations. As shown in the theoretical analysis, LatentAugment is a general model that includes AdvAA and UBS as special cases. Furthermore, the proposed method is simple and computationally efficient in comparison with the existing methods, which requires a generative adversarial network. Experimental results show that the proposed LatentAugment has better performance than previous augmentation methods on the CIFAR-10, CIFAR-100, SVHN, and ImageNet datasets. Finally, an open question remains in the robustness of LatentAugment using the EM algorithm, which typically converges to a local optimum. While we checked the stability of the test accuracy with five runs using different random seeds, the issue of convergence of LatentAugment is an important theme for future research. An application to the object detection, image generation, and text recognition using LatentAugment is also an interesting topic. We leave such directions to future work.

\section{Reproducibility Statement}
Pytorch code of the experiments in this paper can be downloaded from GitHub (\url{https://github.com/KoichiKuriyama/LatentAugment}).

\bibliography{LatentAugment_0503_2023}

\begin{thebibliography}{38}
\providecommand{\natexlab}[1]{#1}
\providecommand{\url}[1]{\texttt{#1}}
\expandafter\ifx\csname urlstyle\endcsname\relax
  \providecommand{\doi}[1]{doi: #1}\else
  \providecommand{\doi}{doi: \begingroup \urlstyle{rm}\Url}\fi

\bibitem[Aitkin \& Aitkin(1996)Aitkin and Aitkin]{aitkin1996hybrid}
Murray Aitkin and Irit Aitkin.
\newblock A hybrid em/gauss-newton algorithm for maximum likelihood in mixture
  distributions.
\newblock \emph{Statistics and Computing}, 6\penalty0 (2):\penalty0 127--130,
  1996.

\bibitem[Cubuk et~al.(2018)Cubuk, Zoph, Mane, Vasudevan, and Le]{cubuk2018auto}
Ekin~D. Cubuk, Barret Zoph, Dandelion Mane, Vijay Vasudevan, and Quoc~V. Le.
\newblock Auto{A}ugment: Learning augmentation policies from data.
\newblock In \emph{CVPR}, 2018.

\bibitem[Cubuk et~al.(2019)Cubuk, Zoph, Shlens, and Le]{cubuk2019rand}
Ekin~D. Cubuk, Barret Zoph, Jonathon Shlens, and Quoc~V. Le.
\newblock Rand{A}ugment: Practical data augmentation with no separate search.
\newblock \emph{arXiv preprint, arXiv:1909.13719}, 2019.

\bibitem[Dempster et~al.(1977)Dempster, Laird, and Rubin]{dempster1977maximum}
Arthur~P Dempster, Nan~M Laird, and Donald~B Rubin.
\newblock Maximum likelihood from incomplete data via the {EM} algorithm.
\newblock \emph{Journal of the Royal Statistical Society: Series B
  (Methodological)}, 39\penalty0 (1):\penalty0 1--22, 1977.

\bibitem[DeVries \& Taylor(2017)DeVries and Taylor]{devries2017improved}
Terrance DeVries and Graham~W. Taylor.
\newblock Improved regularization of convolutional neural networks with cutout.
\newblock \emph{arXiv preprint, arXiv:1708.04552}, 2017.

\bibitem[Gastaldi(2017)]{gastaldi2017shake}
Xavier Gastaldi.
\newblock Shake-shake regularization.
\newblock In \emph{CoRR, abs/1705.07485}, 2017.

\bibitem[Goodfellow et~al.(2014)Goodfellow, Pouget-Abadie, Mirza, Xu,
  Warde-Farley, Ozair, Courville, and Bengio]{goodfellow2014generative}
Ian Goodfellow, Jean Pouget-Abadie, Mehdi Mirza, Bing Xu, David Warde-Farley,
  Sherjil Ozair, Aaron Courville, and Yoshua Bengio.
\newblock Generative adversarial nets.
\newblock \emph{NIPS}, 27, 2014.

\bibitem[Han et~al.(2017)Han, Kim, and Kim]{han2017deep}
Dongyoon Han, Jiwhan Kim, and Junmo Kim.
\newblock Deep pyramidal residual networks.
\newblock In \emph{CVPR}, pp.\  5927--5935, 2017.

\bibitem[Hataya et~al.(2019)Hataya, Zdenek, Yoshizoe, and
  Nakayama]{hataya2019faster}
Ryuichiro Hataya, Jan Zdenek, Kazuki Yoshizoe, and Hideki Nakayama.
\newblock Faster {A}uto{A}ugment: Learning augmentation strategies using
  backpropagation.
\newblock \emph{arXiv preprint, arXiv:1911.06987}, 2019.

\bibitem[He et~al.(2016)He, Zhang, Ren, and Sun]{he2016deep}
Kaiming He, Xiangyu Zhang, Shaoqing Ren, and Jian Sun.
\newblock Deep residual learning for image recognition.
\newblock In \emph{CVPR}, 2016.

\bibitem[Ho et~al.(2019)Ho, Liang, Stoica, Abbeel, and Chen]{ho2019population}
Daniel Ho, Eric Liang, Ion Stoica, Pieter Abbeel, and Xi~Chen.
\newblock Population based augmentation: Efficient learning of augmentation
  policy schedules.
\newblock In \emph{ICML}, 2019.

\bibitem[Inoue(2018)]{inoue2018data}
Hiroshi Inoue.
\newblock Data augmentation by pairing samples for images classification.
\newblock \emph{arXiv preprint, arXiv:1801.02929}, 2018.

\bibitem[Krizhevsky \& Hinton(2009)Krizhevsky and
  Hinton]{krizhevsky2009learning}
Alex Krizhevsky and Geoffrey~E. Hinton.
\newblock \emph{Learning multiple layers of features from tiny images}.
\newblock Technical report, University of Toronto, 2009.

\bibitem[Krizhevsky et~al.(2012)Krizhevsky, Sutskever, and
  Hinton]{krizhevsky2012imagenet}
Alex Krizhevsky, Ilya Sutskever, and Geoffrey~E Hinton.
\newblock Imagenet classification with deep convolutional neural networks.
\newblock \emph{NIPS}, 2012.

\bibitem[LeCun et~al.(1998)LeCun, Bottou, Bengio, and
  Haffner]{lecun1998gradient}
Yann LeCun, L{\'e}on Bottou, Yoshua Bengio, and Patrick Haffner.
\newblock Gradient-based learning applied to document recognition.
\newblock In \emph{Proceedings of the IEEE}, volume~86, pp.\  2278--2324. Ieee,
  1998.

\bibitem[Lemley et~al.(2017)Lemley, Bazrafkan, and Corcoran]{lemley2017smart}
Joseph Lemley, Shabab Bazrafkan, and Peter Corcoran.
\newblock Smart augmentation learning an optimal data augmentation strategy.
\newblock \emph{IEEE Access}, 5:\penalty0 5858--5869, 2017.

\bibitem[Li et~al.(2020)Li, Hu, Wang, Hospedales, Robertson, and
  Yang]{li2020differentiable}
Yonggang Li, Guosheng Hu, Yongtao Wang, Timothy Hospedales, Neil~M Robertson,
  and Yongxin Yang.
\newblock Differentiable automatic data augmentation.
\newblock In \emph{European Conference on Computer Vision}, pp.\  580--595.
  Springer, 2020.

\bibitem[Lim et~al.(2019)Lim, Kim, Kim, Kim, and Kim]{lim2019fast}
Sungbin Lim, Ildoo Kim, Taesup Kim, Chiheon Kim, and Sungwoong Kim.
\newblock Fast {A}uto{A}ugment.
\newblock In \emph{NeurIPS}, 2019.

\bibitem[McLachlan \& Krishnan(2007)McLachlan and
  Krishnan]{mclachlan2007algorithm}
Geoffrey~J McLachlan and Thriyambakam Krishnan.
\newblock \emph{The {EM} algorithm and extensions}, volume 382.
\newblock John Wiley \& Sons, 2007.

\bibitem[Netzer et~al.(2011)Netzer, Wang, Coates, Bissacco, Wu, and
  Ng]{netzer2011reading}
Yuval Netzer, Tao Wang, Adam Coates, Alessandro Bissacco, Bo~Wu, and Andrew~Y.
  Ng.
\newblock Reading digits in natural images with unsupervised feature learning.
\newblock \emph{NIPS Workshop on Deep Learning and Unsupervised Feature
  Learning}, 2011.

\bibitem[Ng et~al.(2012)Ng, Krishnan, and McLachlan]{ng2012algorithm}
Shu~Kay Ng, Thriyambakam Krishnan, and Geoffrey~J McLachlan.
\newblock The {EM} algorithm.
\newblock In \emph{Handbook of computational statistics}, pp.\  139--172.
  Springer, 2012.

\bibitem[Nielsen(2000)]{nielsen2000stochastic}
S{\o}ren~Feodor Nielsen.
\newblock The stochastic {EM} algorithm: estimation and asymptotic results.
\newblock \emph{Bernoulli}, pp.\  457--489, 2000.

\bibitem[Nilsback \& Zisserman(2008)Nilsback and
  Zisserman]{nilsback2008automated}
Maria-Elena Nilsback and Andrew Zisserman.
\newblock Automated flower classification over a large number of classes.
\newblock In \emph{2008 Sixth Indian Conference on Computer Vision, Graphics \&
  Image Processing}, pp.\  722--729. IEEE, 2008.

\bibitem[Russakovsky et~al.(2015)Russakovsky, Deng, Su, Krause, Satheesh, Ma,
  Huang, Karpathy, Khosla, Bernstein, Berg, and
  Fei-Fei]{russakovsky2015imagenet}
Olga Russakovsky, Jia Deng, Hao Su, Jonathan Krause, Sanjeev Satheesh, Sean Ma,
  Zhiheng Huang, Andrej Karpathy, Aditya Khosla, Michael Bernstein,
  Alexander~C. Berg, and Li~Fei-Fei.
\newblock Image{N}et large scale visual recognition challenge.
\newblock In \emph{IJCV}, 2015.

\bibitem[Shorten \& Khoshgoftaar(2019)Shorten and
  Khoshgoftaar]{shorten2019survey}
Connor Shorten and Taghi~M Khoshgoftaar.
\newblock A survey on image data augmentation for deep learning.
\newblock \emph{Journal of Big Data}, 6\penalty0 (1):\penalty0 1--48, 2019.

\bibitem[Tanner(1991)]{tanner1991tools}
Martin~A Tanner.
\newblock \emph{Tools for statistical inference: observed data and data
  augmentation methods}, volume~67 of \emph{Lecture Notes in Statistics}.
\newblock Springer Science \& Business Media, 1991.

\bibitem[Tran et~al.(2017)Tran, Pham, Carneiro, Palmer, and
  Reid]{tran2017bayesian}
Toan Tran, Trung Pham, Gustavo Carneiro, Lyle Palmer, and Ian Reid.
\newblock A bayesian data augmentation approach for learning deep models.
\newblock \emph{arXiv preprint arXiv:1710.10564}, 2017.

\bibitem[Wei \& Tanner(1990)Wei and Tanner]{wei1990monte}
Greg~CG Wei and Martin~A Tanner.
\newblock A {M}onte {C}arlo implementation of the {EM} algorithm and the poor
  man's data augmentation algorithms.
\newblock \emph{Journal of the American statistical Association}, 85\penalty0
  (411):\penalty0 699--704, 1990.

\bibitem[Wu et~al.(2020)Wu, Zhang, Valiant, and R{\'e}]{wu2020generalization}
Sen Wu, Hongyang Zhang, Gregory Valiant, and Christopher R{\'e}.
\newblock On the generalization effects of linear transformations in data
  augmentation.
\newblock In \emph{ICML}, 2020.

\bibitem[Xiao et~al.(2017)Xiao, Rasul, and Vollgraf]{xiao2017fashion}
Han Xiao, Kashif Rasul, and Roland Vollgraf.
\newblock Fashion-mnist: a novel image dataset for benchmarking machine
  learning algorithms.
\newblock \emph{arXiv preprint arXiv:1708.07747}, 2017.

\bibitem[Yamada et~al.(2018)Yamada, Iwamura, and Kisei]{yamada2018shakedrop}
Yoshihiro Yamada, Masakazu Iwamura, and Koichi Kisei.
\newblock Shakedrop regularization.
\newblock In \emph{CoRR, abs/1802.02375}, 2018.

\bibitem[Yun et~al.(2019)Yun, Han, Oh, Chun, Choe, and Yoo]{yun2019cutmix}
Sangdoo Yun, Dongyoon Han, Seong~Joon Oh, Sanghyuk Chun, Junsuk Choe, and
  Youngjoon Yoo.
\newblock Cutmix: Regularization strategy to train strong classifiers with
  localizable features.
\newblock In \emph{Proceedings of the IEEE/CVF International Conference on
  Computer Vision}, pp.\  6023--6032, 2019.

\bibitem[Zagoruyko \& Komodaki(2016)Zagoruyko and Komodaki]{Zagoruyko2016}
Sergey Zagoruyko and Nikos Komodaki.
\newblock Wide residual networks.
\newblock \emph{British Machine Vision Conference}, 2016.

\bibitem[Zhang et~al.(2017)Zhang, Cisse, Dauphin, and
  Lopez-Paz]{zhang2017mixup}
Hongyi Zhang, Moustapha Cisse, Yann~N Dauphin, and David Lopez-Paz.
\newblock Mixup: Beyond empirical risk minimization.
\newblock \emph{arXiv preprint arXiv:1710.09412}, 2017.

\bibitem[Zhang et~al.(2019)Zhang, Wang, Zhang, and Zhong]{zhang2019adversarial}
Xinyu Zhang, Qiang Wang, Jian Zhang, and Zhao Zhong.
\newblock Adversarial {A}uto{A}ugment.
\newblock In \emph{ICLR}, 2019.

\bibitem[Zheng et~al.(2022)Zheng, Zhang, Yan, and Zhang]{zheng2022deepaa}
Yu~Zheng, Zhi Zhang, Shen Yan, and Mi~Zhang.
\newblock Deep {A}uto{A}ugment.
\newblock In \emph{ICLR}, 2022.

\bibitem[Zhou et~al.(2020)Zhou, Li, Xie, Chen, Hong, Sun, and Li]{zhou2020meta}
Fengwei Zhou, Jiawei Li, Chuanlong Xie, Fei Chen, Lanqing Hong, Rui Sun, and
  Zhenguo Li.
\newblock Metaaugment: Sample-aware data augmentation policy learning.
\newblock \emph{arXiv preprint arXiv:2012.12076}, 2020.
\newblock \doi{10.48550/ARXIV.2012.12076}.
\newblock URL \url{https://arxiv.org/abs/2012.12076}.

\bibitem[Zhou et~al.(2021)Zhou, Hong, Hu, Zhou, Ru, Feng, and Li]{zhou2021dha}
Kaichen Zhou, Lanqing Hong, Shoukang Hu, Fengwei Zhou, Binxin Ru, Jiashi Feng,
  and Zhenguo Li.
\newblock Dha: End-to-end joint optimization of data augmentation policy,
  hyper-parameter and architecture.
\newblock \emph{arXiv preprint arXiv:2109.05765}, 2021.
\newblock \doi{10.48550/ARXIV.2109.05765}.
\newblock URL \url{https://arxiv.org/abs/2109.05765}.

\end{thebibliography}
\bibliographystyle{iclr2023_conference}

\appendix
\section{Appendix}
\subsection{Randomly Drawn Subset of the Augmentation Policies}
\label{sec:A1}
Let $\delta_{z}$ be the probability that policy $z$ can be drawn. The
conditional probability with $\delta_{z}$ can be written as:
\[
h_{z}^{\left( t \right)}=\frac{\pi_{z}^{(t)}P\left( y\thinspace\vert\thinspace
{o_{z}\left( x \right),\theta^{\left( t \right)}}\right)
}{\sum\limits_{z\in \mathbb{S}} {\delta_{z}\pi_{z}^{(t)}P\left(
y\thinspace\vert\thinspace {o_{z}\left( x \right),\theta^{\left( t
\right)}}\right) } }.
\]
The expected loss function using the randomly drawn subset given $\delta
_{z}$ is
\[
\begin{split}
\mathcal{E}\left( \Theta \vert \Theta^{(t)},\delta_{z} \right)&=-\mathbb{E}_{\left( x,y
\right)\sim \left( X,Y \right)}\sum\limits_{z\in \mathbb{S}} {\delta_{z}h_{z}^{\left(
t \right)}\log \left( \pi_{z}^{(t)}P\left( y\thinspace\vert\thinspace
{o_{z}\left( x \right),\theta }\right) \right)} \\
&=-\mathbb{E}_{\left( x,y \right)\sim
\left( X,Y \right)}\sum\limits_{z\in \mathbb{S}} {\frac{\delta_{z}\pi_{z}^{(t)}P\left(
y\thinspace\vert\thinspace {o_{z}\left( x \right),\theta^{\left( t
\right)}}\right) }{\sum\limits_{z\in \mathbb{S}} {\delta_{z}\pi_{z}^{(t)}P\left(
y\thinspace\vert\thinspace {o_{z}\left( x \right),\theta^{\left( t
\right)}}\right) } }\log \left( \pi_{z}^{(t)}P\left( y\thinspace\vert\thinspace
{o_{z}\left( x \right),\theta }\right) \right)}.
\end{split}
\]
Assume that the policies of the subset are drawn using simple
random draws: $\delta_{z}=\delta ,\, \forall z\in \mathbb{S}$. Under this assumption, the
expected loss function using a randomly drawn subset is equal to the
expected loss function using the full set:
\[
\mathcal{E}\left( \Theta \vert \Theta^{(t)},\delta \right)=-\mathbb{E}_{\left( x,y \right)\sim
\left( X,Y \right)}\sum\limits_{z\in \mathbb{S}} {\frac{\delta \pi_{z}^{(t)}P\left(
y\thinspace\vert\thinspace {o_{z}\left( x \right),\theta^{\left( t
\right)}}\right) }{\delta \sum\limits_{z\in \mathbb{S}} {\pi_{z}^{(t)}P\left(
y\thinspace\vert\thinspace {o_{z}\left( x \right),\theta^{\left( t
\right)}}\right) } }\log \left( \pi_{z}^{(t)}P\left( y\thinspace\vert\thinspace
{o_{z}\left( x \right),\theta }\right) \right)} =\mathcal{E}\left( \Theta \vert \Theta
^{\left( t \right)} \right).
\]

\subsection{Proof of Theorem \ref{thm:theorem1} (Special Case of LatentAugment)}
\label{sec:A2}

\subsubsection{Uncertainty-based sampling (UBS)}

The loss function of UBS is
\[
\mathcal{L}_{UBS}=\mathbb{E}_{\left( x,y \right)\sim \left( X,Y \right)}\max_{z \in \mathbb{K}}\left[ -\log \left(
P\left( y\thinspace\vert\thinspace {o_{z}\left( x \right),\theta }\right)
\right) \right].
\]
If the inverse scale $\sigma \to 0$, the conditional probability $\Tilde{h}_z$ can be approximated by the indicator function:
\[
\lim_{\sigma \to 0} \Tilde{h}_{z} = 
\lim_{\sigma \to 0} \frac{\exp
\left( -\frac{h_{z}}{\sigma } \right)}{\sum\limits_{r\in \mathbb{K}} \exp \left(
-\frac{h_{r}}{\sigma } \right) }
=I\left[ h_{z}\leq h_{r},\, \forall r\in \mathbb{K} \right],
\]
where $I\left[ \cdot \right]$ is the indicator function such that $I\left[
\cdot \right]=1$ if $h_z \leq h_r,\, \forall r\in \mathbb{K}$, and $I\left[
\cdot \right]=0$, otherwise.

If $\pi_{z}=1
\mathord{\left/ {\vphantom {1 S}} \right. \kern-\nulldelimiterspace} S$ for
all $z$ and $\sigma \to 0$ in the LatentAugment, $\Tilde{h}_{z}\to I\left[ h_{z}\leq h_{r},\,
\forall r\in \mathbb{K} \right]=I\left[ P\left( y\thinspace\vert\thinspace
{o_{z}\left( x \right),\theta^{(t)}}\right) \leq P\left(
y\thinspace\vert\thinspace {o_{r}\left( x \right),\theta^{(t)}}\right) ,\,
\forall r\in \mathbb{K} \right]$. Therefore, the expected loss function of the
LatentAugment is
\[
\begin{split}
\Tilde{\mathcal{E}}\left( \Theta \vert \Theta^{(t)} \right)&\to -\mathbb{E}_{\left( x,y
\right)\sim \left( X,Y \right)}\mathbb{E}_{\mathbb{K}\sim \mathbb{S}}\\
&\left[ \sum\limits_{z\in \mathbb{K}}
{I\left[ P\left( y\thinspace\vert\thinspace {o_{z}\left( x \right),\theta
^{\left( t \right)}}\right) \leq P\left( y\thinspace\vert\thinspace
{o_{r}\left( x \right),\theta^{\left( t \right)}}\right) ,\, \forall r\in \mathbb{K}
\right]\log \left( P\left( y\thinspace\vert\thinspace {o_{z}\left( x
\right),\theta }\right) \right)} +\log \left( 1 \mathord{\left/ {\vphantom
{1 S}} \right. \kern-\nulldelimiterspace} S \right) \right],
\end{split}
\]
\[
\left. \Tilde{\mathcal{E}}\left( \Theta \vert \Theta^{(t)} \right) \right|_{\theta
=\theta^{\left( t \right)}}\to \mathbb{E}_{\left( x,y \right)\sim \left( X,Y
\right)}\mathbb{E}_{\mathbb{K}\sim \mathbb{S}} \max_{z \in \mathbb{K}}\left[ -\log \left( P\left( y\thinspace\vert\thinspace
{o_{z}\left( x \right),\theta^{\left( t \right)}}\right) \right)
\right]-\log \left( 1 \mathord{\left/ {\vphantom {1 S}} \right.
\kern-\nulldelimiterspace} S \right).
\]
Note that the first term is the same as the loss function of the
uncertainty-based sampling evaluated at $\theta =\theta^{\left( t
\right)}$, while the second term is constant. Therefore, $\, \mathrm{\nabla
}_{\theta }\left. \Tilde{\mathcal{E}}\left( \Theta \vert \Theta^{(t)} \right)
\right|_{\theta =\theta^{\left( t \right)}}\to \mathrm{\nabla }_{\theta
}\left. \mathcal{L}_{UBS} \right|_{\theta =\theta^{\left( t \right)}}$, if $\pi
_{z}=1 \mathord{\left/ {\vphantom {1 S}} \right. \kern-\nulldelimiterspace}
S$ for all $z$ and $\sigma \to 0$.

\subsubsection{Adversarial AutoAugment (AdvAA)}

The loss function of AdvAA is
\[
\mathcal{L}_{AdvAA}=-\mathbb{E}_{\left( x,y \right)\sim \left( X,Y \right)}\mathbb{E}_{\mathbb{K}\sim \mathcal{A}\left(
\mathbb{S},\mu \right)}\left[ \frac{1}{K}\sum\limits_{z\in \mathbb{K}} \log \left( P\left(
y\thinspace\vert\thinspace {o_{z}\left( x \right),\theta }\right) \right)
\right],
\]
where $\mathcal{A}\left( \mathbb{S},\mu \right)$ is the adversarial network for the set of
augmentation policies, $\mathbb{S}$ with parameter $\mu $. If $\sigma \to \infty $ in
the LatentAugment, $\Tilde{h}_{z}\to 1 \mathord{\left/ {\vphantom {1 K}}
\right. \kern-\nulldelimiterspace} K$. Assume $\pi_{z}=1 \mathord{\left/
{\vphantom {1 S}} \right. \kern-\nulldelimiterspace} S$ for all $z$ in
LatentAugment. Then, the loss function of the adversarial network with
LatentAugment is equal to the loss function of AdvAA plus constant:
\[
\mathbb{E}_{\mathbb{K}\sim \ \mathcal{A}\left(
\mathbb{S},\mu \right)}\left[ \Tilde{\mathcal{E}}\left( \Theta \vert \Theta
^{\left( t \right)} \right) \right]\to -\mathbb{E}_{\left( x,y \right)\sim \left(
X,Y \right)}\mathbb{E}_{\mathbb{K}\sim \mathcal{A}\left(
\mathbb{S},\mu \right)}\left[
\frac{1}{K}\sum\limits_{z\in \mathbb{K}} \log \left( P\left(
y\thinspace\vert\thinspace {o_{z}\left( x \right),\theta }\right) \right)
\right]-\log \left( 1 \mathord{\left/ {\vphantom {1 S}} \right.
\kern-\nulldelimiterspace} S \right).
\]
Therefore, $\mathrm{\nabla }_{\theta }\mathbb{E}_{\mathbb{K}\sim  \mathcal{A}\left(
\mathbb{S},\mu \right)}\left[
\Tilde{\mathcal{E}}\left( \Theta \vert \Theta^{\left( t \right)} \right) \right]\to
\mathrm{\nabla }_{\theta }\mathcal{L}_{AdvAA}$, if $\pi_{z}=1 \mathord{\left/
{\vphantom {1 S}} \right. \kern-\nulldelimiterspace} S$ for all $z$ and $\sigma
\to \infty $.

\subsection{Computer Resources}
\label{sec:A3}
We train the models with the LatnetAugment using computers with 4 NVIDIA RTX 2080Ti GPUs and 64 GB memory.
 
\subsection{Hyperparameters}
\label{sec:A4}
\begin{table}[htbp]
\caption{Hyperparameters for the experiment. LR represents the learning
rate, whereas WD represents the weight decay.}
\label{tab_a1}
\vskip 0.15in
\begin{center}
\begin{tabular}{cccccc}
\toprule
Dataset &
Model &
LR &
WD &
Batch Size &
Epoch \\
\midrule
CIFAR-10 &
Wide-ResNet-40-2 &
0.1&
0.0002&
128&
200 \\

CIFAR-10 &
Wide-ResNet-28-10 &
0.1&
0.0005&
128&
200 \\

CIFAR-10 &
Shake-Shake (26 2x32d) &
0.01&
0.001&
128&
1800 \\

CIFAR-10 &
Shake-Shake (26 2x96d) &
0.01&
0.001&
128&
1800 \\

CIFAR-10 &
Shake-Shake (26 2x112d) &
0.01&
0.001&
128&
1800 \\

CIFAR-10 &
PyramidNet$+$ShakeDrop &
0.05&
5E-05&
64&
1800 \\

CIFAR-100 &
Wide-ResNet-40-2 &
0.1&
0.0002&
128&
200 \\

CIFAR-100 &
Wide-ResNet-28-10 &
0.1&
0.0005&
128&
200 \\

CIFAR-100 &
Shake-Shake (26 2x96d) &
0.01&
0.0025&
128&
1800 \\

SVHN&
Wide-ResNet-28-10 &
0.005&
0.001&
128&
160 \\

ImageNet&
Resnet-50&
0.1&
0.0001&
256&
270 \\
\bottomrule
\end{tabular}
\label{tab4}
\end{center}
\end{table}

\subsection{Test accuracies using the same transformations as AA and AdvAA.}
This section provides comparison between proposed LatentAugment (LA) and AutoAugment (AA) 
or Adversarial AutoAugment (AdvAA) using same subset size and transformations. 
To compare with the AA, we tested the proposed model
using the same transformations as AA, which uses the policy set with
SamplePairing \citep{inoue2018data}, instead of Mixup \citep{zhang2017mixup}, and
finally applied the Cutout, instead of Cutmix. Table \ref{tab_acc_same_aa} shows  
the test accuracies for the Wide-ResNet 40-2 model and Wide-ResNet 28-10 of
the LA using the same transformation as AA. This table also provides the result of 
UBS using same transformations as AA, reported by \citet{wu2020generalization}.

AdvAA applies the same transformations as AA, although the subset size of AdvAA is 8. 
To compare with AdvAA, we tested the model using the same subset size. 
Table \ref{tab_acc_same_advaa} shows the test accuracy of Wide-ResNet 28-10 of the 
LA using the same transformations as AA with subset size $K=8$.

\begin{table}[htbp]
\begin{center}
\caption{Test accuracies (\%) on CIFAR-10 using the same transformations as AutoAugment (AA). On the proposed LatentAugment (LA), averages of five runs are reported.}
\vskip 0.15in
\begin{tabular}{cccc}
\toprule
Model & AA & UBS & LA ($K=6$) \\ 
\midrule
WRN40-2 &96.30 & - & 96.91$\pm$0.05 \\
WRN28-10 & 97.32 & 97.75 & 98.01$\pm$0.05 \\
\bottomrule
\end{tabular}
\label{tab_acc_same_aa}
\end{center}
\end{table}

\begin{table}[htbp]
\begin{center}
\caption{Test accuracies (\%) on CIFAR-10 using the same subset size and transformations as Adversarial AutoAugment (AdvAA). On the proposed LatentAugment (LA), an average of five runs is reported.}
\vskip 0.15in
\begin{tabular}{ccc}
\toprule
Model & AdvAA & LA ($K=8$) \\ 
\midrule
WRN28-10 & 98.10 & 98.16$\pm$0.07 \\
\bottomrule
\end{tabular}
\label{tab_acc_same_advaa}
\end{center}
\end{table}

\subsection{Transferability across datasets and architectures}
This section evaluates transferability with LatentAugment across different datasets and model architectures. We first take a snapshot of the unconditional probability $\pi_z$ of ResNet-50 on ImageNet using LA, and then apply the fixed $\pi_z$ to train the models of Wide-ResNet 40-2 on CIFAR-10 or CIFAR-100 using LA. Table \ref{tab_a2} provides the experimental result of transferability. It suggests that LA with policy transfer has still good performance.

\begin{table}[htbp]
\begin{center}
\caption{The test accuracies of the transfer the unconditional probability of augmentation policies.}
\vskip 0.15in
\begin{tabular}{cccc}
\toprule
Dataset  & Baseline     & LA (direct)  & LA (policy transfer)  \\
\midrule
CIFAR-10  & 94.7     & 97.27$\pm$0.09  & 97.21$\pm$0.09  \\

CIFAR-100 & 74.0     & 80.90$\pm$0.15 & 80.77$\pm$0.14 \\
\bottomrule
\end{tabular}
\label{tab_a2}
\end{center}
\end{table}

\subsection{Additional experiments using other datasets}
This section provides the results of additional experiments using datasets of MNIST \citep{lecun1998gradient}, Fashion MNIST \citep{xiao2017fashion}, and Oxford flowers102 \citep{nilsback2008automated}.

\subsubsection{MNIST}
The MNIST is a large database of handwritten digits. It has a total of 70,000 images, including 60,000 for the training set and 10,000 for the test set.  Each example is a 28x28 grayscale image, associated with a label from 10 classes.  
The baseline is trained with standard data augmentation using zero-padding, and random crops. The proposed LatentAugment first applies the baseline preprocessing, then applies LatentAugment using six policies randomly drawn from 256 policies, and finally applies the Cutout. We use hyperparameters of WRN40-2 on CIFAR-10 in Table \ref{tab4}.

\subsubsection{Fashion MNIST}
The Fashion MNIST is a dataset of Zalando's article images—consisting of a training set of 60,000 examples and a test set of 10,000 examples. Each example is a 28x28 grayscale image, associated with a label from 10 classes. 
The baseline is trained with standard data augmentation using horizontal flips with 50{\%} probability, zero-padding, and random crops. The proposed LatentAugment first applies the baseline preprocessing, then applies LatentAugment using six policies randomly drawn from 256 policies, and finally applies the Cutmix. We use hyperparameters of WRN40-2 on CIFAR-10 in Table \ref{tab4}

\subsubsection{Oxford flowers102}
Oxford 102 Flower is an image classification dataset consisting of 102 flower categories. The flowers chosen to be flower commonly occurring in the United Kingdom. Each class consists of between 40 and 258 images. Baseline augmentation uses the standard
Inception-style pre-processing, including horizontal flips with 50{\%}
probability and random distortions of colors. The proposed LatentAugment
first applies the baseline preprocessing, then applies LatentAugment using
six policies randomly drawn from 256 policies, and finally applies the
Cutmix. We use hyperparameters of ResNet-50 on ImageNet in Table \ref{tab4}

\begin{table}[htbp]
\begin{center}
\caption{The test accuracies of the additional experiments using MNIST \citep{lecun1998gradient}, Fashion MNIST \citep{xiao2017fashion}, and Oxford flowers102 \citep{nilsback2008automated}. }
\vskip 0.15in
\begin{tabular}{cccc}
\toprule
Dataset  & Model & Baseline     & LA  \\
\midrule
MNIST  & WRN40-2 & 99.66     & 99.77$\pm$0.02    \\
Fashion MNIST & WRN40-2 & 94.54     & 96.04$\pm$0.06  \\
Flowers102 & ResNet-50 & 95.70     & 98.19$\pm$0.09  \\
\bottomrule
\end{tabular}
\label{tab_a3}
\end{center}
\end{table}

\subsection{Convergence of Loss Functions}
\label{sec:A5}
\begin{figure}[htbp]
\centerline{\includegraphics[width=6in]{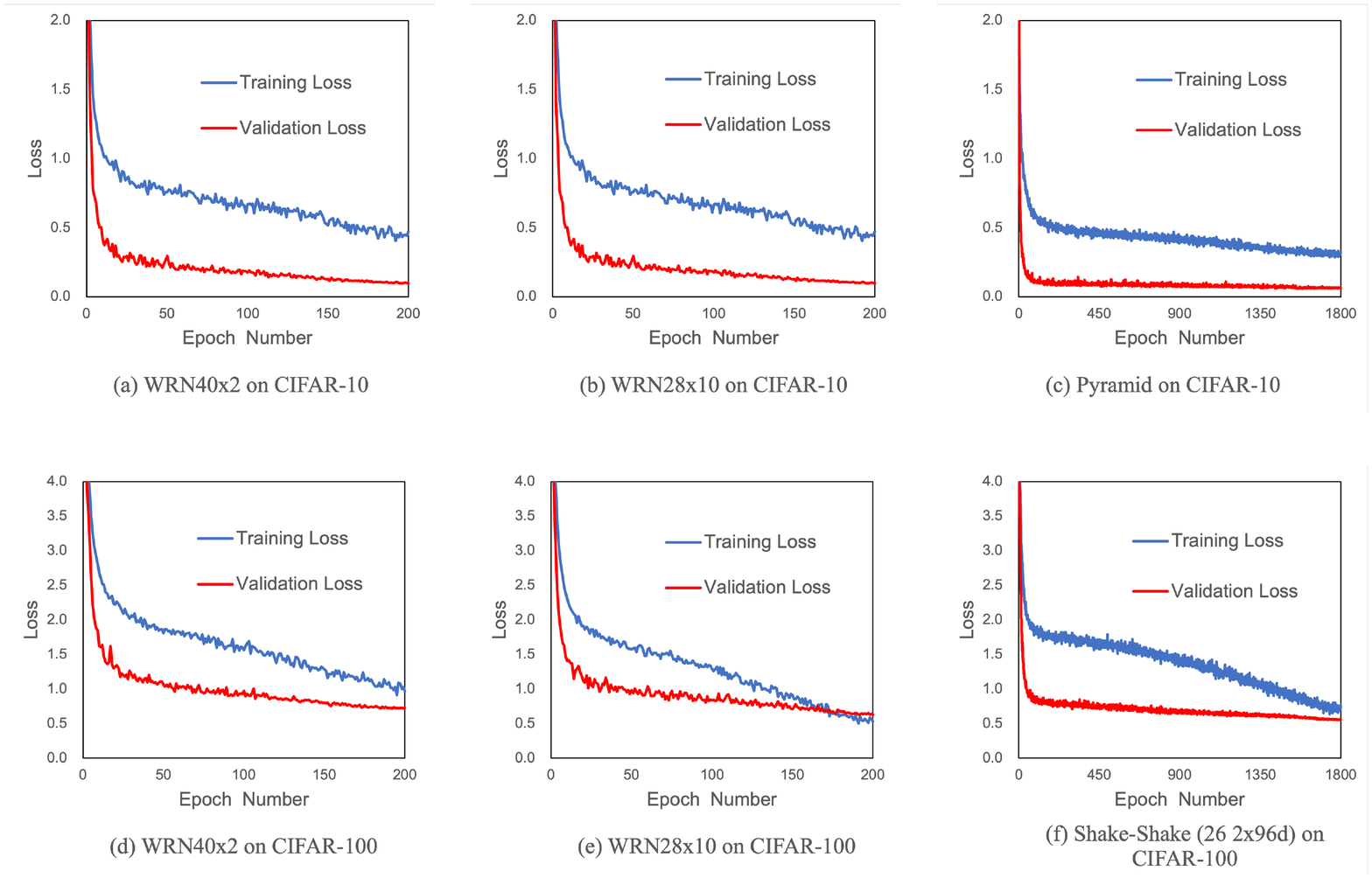}}
\caption{The loss functions of the different network models on CIFAR-10 and CIFAR-100.}
\label{FigA2}
\end{figure}

Convergence of the EM algorithm is usually defined as a sufficiently small change in the loss function \citep{aitkin1996hybrid}. To confirm the convergence, Figure \ref{FigA2} shows the loss functions using the network models of Wide-ResNet 40-2, Wide-ResNet 28-10, Pyramid, and Shake-Shake on CIFAR-10 and CIFAR-100. This figure indicates the convergence of an estimation using the EM algorithm.

\end{document}